\documentclass[sigconf, anonymous=false]{acmart}

\renewcommand\footnotetextcopyrightpermission[1]{} 
\usepackage{booktabs, colortbl} 
\usepackage{siunitx}
\usepackage{multirow}
\usepackage{graphicx}
\usepackage{amsmath}
\usepackage[pagewise]{lineno}
\usepackage{natbib}
\setcitestyle{numbers}
\usepackage{url}
\usepackage[multiple]{footmisc}

\newcommand{\RNum}[1]{\lowercase\expandafter{\romannumeral #1\relax}}

\setcopyright{none}

\begin{document}
\title[Using Active Learning Methods to Strategically Select Essays for Automated Scoring]{Using Active Learning Methods to Strategically Select Essays for Automated Scoring}

\author{Tahereh Firoozi}
\affiliation{%
  \institution{Center for Research in Applied Measurement and Evaluation,}
  \institution{University of Alberta}
  \city{Edmonton}
  \state{Canada}
}
\email{tahereh.firoozi@ualberta.ca}

\author{Hamid Mohammadi}
\orcid{0000-0002-8854-9342}
\affiliation{%
  \institution{Computer Engineering Department,}
  \institution{Amirkabir University of Technology}
  \city{Tehran}
  \state{Iran}
}
\email{hamid.mohammadi@aut.ac.ir}

\author{Mark J. Gierl}
\affiliation{%
  \institution{Center for Research in Applied Measurement and Evaluation}
  \institution{University of Alberta}
  \city{Edmonton}
  \state{Canada}
}
\email{mark.gierl@ualberta.ca}

\renewcommand{\shortauthors}{T. Firoozi et al.}

\begin{abstract}
Research on automated essay scoring has become increasing important because it serves as a method for evaluating students’ written-responses at scale. Scalable methods for scoring written responses are needed as students migrate to online learning environments resulting in the need to evaluate large numbers of written-response assessments. The purpose of this study is to describe and evaluate three active learning methods than can be used to minimize the number of essays that must be scored by human raters while still providing the data needed to train a modern automated essay scoring system. The three active learning methods are the uncertainty-based, the topological-based, and the hybrid method. These three methods were used to select essays included as part of the Automated Student Assessment Prize competition that were then classified using a scoring model that was training with the bidirectional encoder representations from transformer language model. All three active learning methods produced strong results, with the topological-based method producing the most efficient classification. Growth rate accuracy was also evaluated. The active learning methods produced different levels of efficiency under different sample size allocations but, overall, all three methods were highly efficient and produced classifications that were similar to one another.
\end{abstract}


\maketitle

{

An extraordinary range of learning opportunities are now available through the use of instructional technologies that permit students to access massive open online courses (MOOC) and other online learning programs. For example, the World Economic Forum claimed that 21 million students registered for Coursera’s online courses in 2016. The pandemic only served as a catalyst for the migration to online learning as the number of registrations skyrocketed. Coursera enrollment increased more than three-fold bringing the figure to 71 million in 2020, with an additional 21 million registrations in 2021 bring the most recent count to 92 million \citep{nam2022world} that students have abundant opportunities to access online learning resources—and that they are capitalizing on these opportunities. One important challenge that remains to be addressed in the world of online teaching and learning focuses on the development and administration of educational assessments. In particular, administering written-response assessments that yield valid and reliable test score interpretations poses an important challenge because of the scoring process. A written-response assessment such as an essay must be evaluated by a rater in order to yield inferences about the student’s knowledge, skills, and competencies. The traditional method for scoring involves training a human rater to interpret and apply a rubric that can be used to score the students’ responses.

Unfortunately, human scoring is time consuming because it requires human raters to evaluate a large number of essays. It is also expensive because it requires extensive logistical efforts to hire human raters, train the raters to consistently interpret and apply the scoring rubric, and deploy these raters to evaluate each student’s written-response task. But an even more important challenge exists. Using human raters to evaluate written-response assessments is virtually impossible to scale in a timely and cost-effective manner. If, for instance, 100 students complete a written-response assessment such as an essay, then 100 essays must be scored by the human raters. This scale is reasonable with enough trained raters. If 92 million Coursera students complete an essay as part of their online courses, then 92 million essays must be evaluated by raters. This scale is not reasonable because the time and expense required to train a legion of human raters who must then score the essays is prohibitive.

On way to address this scaling challenge is to implement automated essay scoring (AES) so that machines can be used to help humans score students’ written-response assessments. AES can be described as the use of computer algorithms to score unconstrained open-ended written tasks by having a computer mimic the human raters \cite{hermis2014state,shermis2010automated}. Research on the development and application of AES has become increasing important in the last decade as practitioners attempt to implement methods that can be used to efficiently and accurately scoring students’ written-responses at scale. The need for these methods has only been amplified in the past three years as students migrate to online learning environments en masse resulting in the need for new scoring practices that allow educators to evaluate large numbers of written-response assessments efficiently and economically.

\section{Automated Essay Scoring: A Description of the Practice and the Process}
\label{automated}

AES can be described as the practice of evaluating and scoring written text using computer programs. An AES system is a computer program that is designed to evaluate student responses so that the program yields scores that are similar to those of trained human raters \citep{shermis2014state}. AES is a statistical classification method where input linguistic features in the text are mapped to well-defined output, such as an essay score, so that the input and output can be statistically related to one another. This mapping function, called a scoring model, is then used to classify any future instance of the input text onto the output score. The use of a scoring model allows educators to scale the assessment because, instead of a human, the computer can be used to score students’ written tasks. To emulate human scoring, the AES program builds the scoring model using a range of techniques drawn from natural language processing and computational linguistics where l features are extracted from the example instances, called the training dataset, that have been scored by human raters. When a training dataset is used, the AES method is said to be supervised. A supervised machine learning algorithm uses scored training samples from human raters on a specific written-response prompt. The algorithm learns the behaviour of the raters by analyzing the classification patterns in the scored essays. This step yields statistical weights that, in turn, can be used in the scoring model step so that a new set of written-response tasks can be classified meaning that the model can then be used to score the written responses using the same essay but with a different group of students. While the term “essay” is included in the phrase AES, many different types of written-response tasks can be scored using this method, ranging from short-answer response to long essays. In this study, all of these potential written-response assessments will simply be referred to as essays.

The AES process can be described in four steps. The first step is pre-processed. Pre-processing requires the student response data to be available in electronic form so that it can be formatted and cleaned. In an online learning system, students’ data are available immediately in an electronic form. Transformations are then conducted so that, for instance, raw numeric scores from human raters are annotated into text scoring labels that can be read by an AES system. After formatting and cleaning is completed, the students’ written responses are converted into a form that is readable by the AES system.

The second step is feature extraction. Feature extraction is a process of objectively transcribing the input text into different characteristics or features that can be used by the machine learning algorithm to represent the text. Traditional AES approaches focus on constructing and extracting discriminating linguistic features from the text that could be used as variables in order to predict the final essay score \citep{attali2013validity, mcnamara2015hierarchical, mcnamara2014automated, page1994computer}. The benefit of the traditional approaches is that the linguistic features are identified prior to the AES analysis thereby providing interpretable indicators of written-response quality. The drawback of the traditional approaches is that the predictive performance may not reach a high level of accuracy \citep{shin2021more}.

To overcome the limitations of the traditional approaches, alternative AES approaches can be used that attempt to maximized the predictive accuracy of reproducing the final essay score. These modern AES approaches automatically detect and extract features that can be used to model the association between each essay and the final essay score with the goal of maximize the predictive accuracy of the scoring model \citep{dong2017attention, kim2014convolutional}. For instance, deep learning is a modern feature extraction method designed to maximize the predictive accuracy of the scoring model. A traditional AES approach starts with relevant features being manually extracted from the text. These features are used to create a model that categorizes important text features. A modern AES approach using deep learning starts with relevant features being automatically extracted from the text. The extraction process requires end-to-end learning which means that a neural network is given a text and the data with the rater’s scores. The task for the network is to learn how to reproduce the scores as output using the text as input. The word “deep” refers to the number of hidden layers in the neural network. Traditional neural networks may contain 2-3 hidden layers while deep neural networks can have as many as 150. This large number of layers help the network learn minute details that, in turn, are used to maximize the score prediction. The strength of modern AES approach is that the model can predict the essay scores with high accuracy. The weakness of a modern AES approach is that the complex feature structures used to maximize the predictive accuracy of the scoring model are challenging to interpret linguistically. In other words, the modern approach produces highly predictive results using variables that are often challenging to interpret as a meaningful set of language variables. In addition, the modern approaches require larger essay samples sizes compared to the traditional approaches.

The third step is the creation of a scoring model using machine learning algorithms. A scoring model contains a list of the extracted features from the second step. This model contains the input that are then mapped onto the human rater scores that serve as the output so that a well-defined relationship exists for transforming the input to the output. Machine learning algorithms are given the task of learning the relationship between the input text features and output essay scores by analyzing sample responses in the training dataset in order to learn the classification process (i.e., supervised machine learning). Many different algorithms can be used to learn the classification process. Machine learning is an evidence-based process where the input features are mapped onto the output scores with the goal of developing a text classifier capable of accurately scoring students’ written responses. Because the mapping function required to produce this transformation is not immediately apparent, the algorithm must learn how best to describe the function by analyzing human rater data in the training dataset. 

The fourth step is essay scoring. The machine learning model is used to score written responses using the same essay prompt but with a different group of students called the validation sample. When the AES system creates a model that can be used to score data from one existing written-response or essay prompt, the model is said to be prompt specific. When the AES system creates a model that can be used to score data from a collection of prompts that are designed to be interchangeable, the model is said to generic. The majority of AES studies have focused on the performance of prompt-specific scoring models because they yield more accurate score predictions.

After the written-response scores are classified (i.e., scored or graded), the accuracy of the scoring model can be evaluated using different performance measures. Model validation in AES often depends on comparing the similarity between the model performance and human raters \citep{attali2013validity,chung2003issues,williamson2012framework}. In this comparison, human judges are considered the “gold standard” and function as the explicit criterion for evaluating the performance of AES scoring model. Various validity coefficients have been adopted as evaluation metrics to measure agreement. Three commonly used measure of score agreement are exact-agreement percentage, kappa, and quadratic-weighted kappa. Exact-agreement percentage is the exact matching between human and computer scores. It is reported as a percentage. Kappa is a measure of agreement that takes into consideration agreement by chance alone. Kappa provides a chance-corrected index and is based on the ratio of the proportion of times the agreement is observed to the maximum proportion of times that the agreement is made while correcting for chance agreement \citep{siegelnj}. It ranges from one, when agreement is perfect, to zero when agreement is not significantly better than chance. A kappa of 1.0 indicates perfect agreement between the human rater and computer whereas a kappa of 0.0 indicates the agreement is equivalent to only a chance or random outcome. Kappa, however, does not account for the degree of disagreement. Therefore, a weighted kappa score called quadratic weighted kappa (QWK) is used to address this limitation. In QWK, $i$ represents a human-rated score, $j$ represents a machine-rated score, and $N$ is the number of possible ratings. A weight matrix $W$ can then be constructed as follows:

\begin{equation}
\label{eq:selfattention}
\resizebox{.3\hsize}{!}
{
$W_{i,j} = \frac{(i-j)^2}{(N-1)^2}$
}
\end{equation}\\

Next, a matrix $O$ is created where $O_{i,j}$ represents the number of essays that receive a rating $i$ by the human and a rating $j$ by the machine. An expected count matrix $E$ is computed as the outer product of histogram vectors of the two ratings. The matrix is normalized so that the sum of elements in $E$ and $O$ are the same. $QWK$ can therefore be calculated as follows:

\begin{equation}
\label{eq:selfattention}
\resizebox{.5\hsize}{!}
{
$QWK = 1 - \frac{\sum_{i,j}^{} W_{i,j} O_{i,j}}{\sum_{i,j}^{} W_{i,j} E_{i,j}}$
}
\end{equation}\\

A $QWK$ of 0.80 or higher is considered to indicate strong agreement.

\section{The Cost of Using AES}
\label{cost}

AES offers educators many important benefits for scoring students’ written-response assessments in online learning environments. AES systems yields scores that consistently agree with those obtained from human raters. In fact, many studies have even demonstrated that AES systems can classify scores at a rate as high, if not higher, that the agreement among human raters themselves. AES is a scalable method thereby allowing educators to evaluate large numbers of written-response assessments efficiently and economically. AES has broad applicability, as it can be used for formative-based low-stakes assessment as well as summative-based high-stakes testing. It is also extraordinary fast requiring just seconds to score thousands of written-response tasks. AES has been used to score essays and constructed-response tasks in a variety of writing genres including persuasive, descriptive, narrative, cause-and-effect, expository, comparative, problem-and-solution, argumentative, response to issue, and response to literature. New studies are also being conducted to demonstrate on how to score essays in low-resources languages other than English such as Persian (e.g., Firoozi \& Gierl, 2022).

But AES also comes with one significant cost. Most AES systems that are used with written-response assessments in education are supervised which means that training data are required to develop the scoring models. Currently, the best way to ensure high agreement between the human rater and computer is to calibrate the system with a large number of scored tasks that represent a wide range of score levels. In other words, more training data are better for supervised machine learning and the data should be representative of all score levels. In addition, the application of deep learning models—which serve as the current state-of-the-art for AES—are preferable because they predict scores with a high level of accuracy. But as we noted earlier, these modern approaches require larger samples of essays.

This requirement of providing the AES system with large numbers of scored essays that represent a full range of performance levels is challenging to address, particular in the world of online education. In most cases, the starting point for conducting AES using written-response assessments begins with accessing a large set of essays. This type of data is easy to access in an online educational platform. Unfortunately, these readily accessible essays are both unstructured and unscored. The challenge then becomes securing enough scored data in order to implement the four-step AES process. In most cases, it is problematic to develop a reliable AES system for scoring written assessments because it is difficult to collect enough scored data that can be used to train the system. The difficulty stems from asking human raters to score a large number of essays in a short period of time.

The purpose of this study is to address this problem. Active learning (AL) methods allows educators to build modern AES systems while adhering to a limited scoring budget because the number of essays that must be scored by human raters is minimized. We describe and evaluate three AL methods that can be used to substantially minimize the number of essays that must be scored by human raters while still providing the data needed to efficiently train a modern AES system based on deep features.

\section{Active Learning Methods}
\label{active}

AL is a branch of artificial intelligence that attempts to use a minimal set of labeled data to build a robust machine learning model \citep{settles2012active}. AL is capable of solving problems in those situations where unlabeled data may be abundant but annotating this data is a slow and expensive process. In the context of AES, unlabeled data refers to the essays and labelling data through annotation refers to scoring essays using human raters. We will use the terms essays that are not scored by human raters and essays that are scored by human scoring as unlabeled and labeled data, respectively, throughout this study. Using an AL approach, the machine learning system is first queried to determine regions of classification inconsistency. Next instances in that region are sampled and scored by human raters. Then the newly scored instances are added to the training data. Finally, the classifier is retrained using the updated training data. This process is repeated until a final scoring model is produced. This process helps the machine to achieve high levels of accuracy using as little human scored data as possible. Different AL approaches can be used including query synthesis, stream-based selective sampling, and pool-based sampling \citep{trajkova2021active}. This study focuses on pool-based sampling because in the context of AES there is a small number of scored essays (i.e., labelled data) and a large number of unscored essays (i.e., unlabeled data) that can be identified within a closed (i.e., static or non-changing) system \citep{liu2020unsupervised}. Hence the process of sampling data points from a pool of essays serves as a well-defined task because the total number of essays in the pool can be specified.

One of the most widely used pool-based sampling methods is uncertainty sampling \citep{he2019towards}. With this method, the algorithm explores the sample of data points where the model is least confident about identifying the correct score. The level of confidence for the classification of each data point is determined based on the extent to which that data point is distant from the margin of the decision boundary in the classification task \citep{nguyen2022measure}. The prediction error of the model is higher for the data points that lie somewhere in the transition zone from one score to another score. Essays that are found in this zone are flagged and used in the training dataset because the rater’s score helps resolve the uncertainty associated with classifying the essays in this transition zone.

The uncertainty method has been used to solve data sparsity problems in educational testing. For instance, \cite{horbach2016investigating} used AL methods to score short-answer prompts using data from the Automated Student Assessment Prize (ASAP) competition. The ASAP competition was organized by Kaggle and sponsored by the Hewlett Foundation in 2012. The competition focused on the application and effectiveness of AES technology as it applies to essay scoring\cite{shermis2014state}. The task in the competition was for the AES systems to reproduce the essay scores initially produced by human raters. Scores from human raters were obtained on 12,978 written for eight prompts (four traditional writing genres and four source-based writing genres) taken from students in six US states at three grade levels (Grades 7, 8, and 10) who wrote their exams under standardized testing conditions. The essays were written by an ethnically diverse, gender-balanced sample of students and graded by trained teacher raters using eight scoring rubrics (five holistic scoring rubrics; one two-trait rubric; two multiple-trait rubrics). Horbach and Palmer demonstrated that the uncertainty method could be used for predicting essay scores. Their AES model produced kappa estimates using only 300 scored essays that were comparable to the kappa produced using the full sample of ASAP essays. However, the performance of the AL methods was not consistent across all the eight essay prompts in the ASAP dataset. There was a consistent improvement for four prompts using the uncertainty methods while the remaining four prompts showed no improvement.

\cite{hastings2018active} used uncertainty sampling to address the cost of essay scoring when attempting to provide feedback on students’ explanatory essays in an intelligent tutoring system. Results of their study showed that the uncertainty method reduced the need for large training dataset while still achieving high scoring accuracy. More specifically, when they used the essays identified by the uncertainty method—which represented 30\% of the total sample—they could achieve almost the same agreement accuracy when compared to the results produced using the entire dataset.

\cite{dronen2015effective} investigated the performance of two pool-based AL methods for AES, uncertainty sampling and topological-based AL. Topological-based AL focuses on resembling each class cluster using a minimal set of data points. The model learns the general shape and position of the cluster for each class in the feature space from the selected data points. The idea of shape comes from the relative distance or closeness of data points in the feature space. Hence, topological-based AL algorithms are formed by calculating the degree of similarity between the data points. Each data point is a feature vector (i.e., a vector containing a number of essay features) that can be used to numerically represent an essay in a feature space. \cite{dronen2015effective} trained a least square regression model using the entire ASAP dataset and then tested the two AL models on a simulated dataset that was based on the same ASAP features. Results from their study showed that in almost all the prompts in their simulated dataset, training with 30-50 essays that were identified using the AL algorithms provided approximately the same performance as using a model that was trained using the full dataset.

One of the limitations of \citeauthor{dronen2015effective} (2015) study was that the AL experiments were conducted using the entire training dataset which is not computationally efficient. To address this problem, \citeauthor{hellman2019scaling} (2019) introduced and evaluated batch-mode AL algorithms to build a cost-efficient AES scoring model for their web-based writing software in order to provide students with feedback on their writing assignments. Batch-mode AL is a practical technique where the most informative essays are identified in each training iteration. Batch-mode AL selection serves as an improvement over single instance selection because by sequentially selecting a single essay in each training iteration, a set of essays can be selected over all of the iterations \citep{lourentzou2018exploring}. The general workflow for batch-model AL begins with a given set of training data that contain scores where a model is built and fit to the training data. Then, a set of candidate data points is chosen based on a sampling technique (e.g., uncertainty or topological based), and the accuracy of the model is evaluated at the locations of the candidate data points. Next, each candidate data point is assigned a score based on a machine learning scoring function. Finally, the candidates with the highest scores are selected, human scores are added to the training data, and the new model is evaluated \citep{maljovec2014topology}. \citeauthor{hellman2019scaling} (2019) used uncertainty and topology-based selection for sampling data points at each batch. Results of their study showed that the AL algorithms could be used to achieve 95\% of the performance produced using the full sample across all of the ASAP essay prompts.

\section{Transformer-Based Modeling and Active Learning As Applied to AES}
\label{transformer}

The studies conducted by \citeauthor{horbach2016investigating} (2016), \citeauthor{hastings2018active} (2018), \citeauthor{dronen2015effective} (2015), and \citeauthor{hellman2019scaling} (2019) demonstrated how AL methods can be used to decrease the number of scored essays required to create a reliable AES system. All four studies also used traditional machine learning models containing handcrafted linguistic features (i.e., the traditional approach as described in step 2 of our AES overview). However, as we noted earlier, increasing the number of features can dramatically improve the accuracy of score prediction \citealp{uto2021review}. Moreover, recent advancements in NLP now allow researchers to conceptualize written-response tasks, such as student essays, as an information-rich vector representation. In particular, the use of transformers such as the Bidirectional Encoder Representations from Transformers model \citep{devlin2018bert}, also known as BERT, serves as a deep learning language framework where every input feature is connected to every output feature meaning that the linguistic context of the input can be included in this vector representation for a written response. For example, the word “bank” has different meanings in these contexts: “Bank of Canada”, “memory banks”, and “river bank”. When the context of the word in not considered, the numerical representation (i.e., feature vectors) might not be calculated accurately which leads to the misrepresentations of the essays in feature space. These misrepresentations negatively affect the performance of AL algorithms because essay sampling is typically based on either the position (uncertainty sampling) or the shape (topology-based algorithms) of the essays in feature space. 

To address this misrepresentation problem, text as unstructured data needs to be represented in an information-rich numerical presentation processed with a context by the machine learning model. The numerical representation consists of a list of features forming a vector. Transformers models such as BERT are tools to generate text representation that can produce feature vectors containing the information required to conduct different NLP tasks. The power of BERT comes from its ability to learn a language model from a large corpus of plain text. Because the context of the language is considered, the model thoroughly “understands” how a language is structured as well as the relationships between the language components (e.g., words, sentences, paragraphs). BERT, as a deep learning language model, is trained by completing a series of textual predictive problems which allows the model to learn the language context \citep{prabhu2022hybrid}.

Deep learning language models such as BERT must also be fine-tuned in order to accurately model the context of a language. Fine tuning is the process of training a neural network on a specific application to enrich the knowledge and provide the context required by the neural network to model a particular topic \citep{kong2022hierarchical}. In the current study, we use the complete ASAP dataset to demonstrate how AL methods can reduce the total number of essays required to produce a reliable AES system. As a result, BERT was fine-tuned on the ASAP dataset in our study in order to improve its ability to score the essays. Fine tuning is an essential step when using language models because it provides language context which can enhance the performance of the model \citep{prottasha2022transfer}. The baseline BERT model is first used for the task of classifying texts and scoring essays using the ASAP dataset. The baseline model learns the properties of texts with different essay scores and incorporates this information into the generated feature vectors. The fine-tuned BERT model is then used to classify and score unscored essays in the ASAP dataset. The fine-tuned model learns the properties of texts with different essay scores and incorporates this information into the generated feature vectors which now contains the language context. A feature vector with language context is much more discriminating and as a result performs better at classifying texts and scoring essays. As an example, \autoref{tab:bertaccuracy} compares the BERT language model’s classification accuracy on the ASAP dataset before and after fine-tuning. QWK increases noticeably after the context is added to the feature vectors.

\begin{table}[!htbp]
\centering
\caption{BERT Accuracy Using ASAP Data Before and After Fine Tuning}
\label{tab:bertaccuracy}
\begin{tabular}{cc}
\toprule
\textbf{Model} & \textbf{QWK} \\
\midrule
BERT before fine-tuning & 0.61 \\
BERT after fine-tuning & 0.78 \\
\bottomrule
\end{tabular}
\end{table}

\section{Minimizing the Scoring Task for Human Raters: A Review of Three AL Methods}
\label{minimizing}

The time and cost of human scoring can be decreased by reducing the number of essays that must be evaluated as training set for scoring machine. By implementing an AL strategy, it is possible to select a relatively small set of essays that will provide the most useful information for creating a reliable AES scoring model. In this study, we describe and evaluate three different AL methods: the uncertainty-based method, the topological-based method, and a hybrid method which combines properties from both the uncertainty- and topological-based methods.

\subsection{Uncertainty-Based Method}

Selecting the data points that could yield the most information to the machine learning model is the basis of uncertainty-based AL. These data points are usually ones where a discriminative machine learning model has the lowest confidence associated with a classification \citep{mai2022effectiveness}. In other words, the predicted probability that the data points belong to a specific class is not high enough to confidently select one of the classes as the affiliated class. The most uncertain essays are usually located at the boundary of class clusters in the feature space where the data points from two or more classes are present. As a result, these regions could not be assigned to a single class with a high degree of certainty.

Neural networks predict the association probabilities as a probability distribution over the classes. The expected association probability or confidence is used as a measure of the neural network’s uncertainty \citep{nguyen2022measure}. The fine-tuned BERT classifier is then used to calculate the uncertainty for the unscored essays. Fine tuning the BERT classifier on the data points with the highest predicted uncertainty increases the confidence in the prediction when using unscored essays. \autoref{fig:figure_1} demonstrates how data points are selected using the uncertainty-based method using the ASAP dataset. The black points are the essays. The green points are the essays selected for scoring using the uncertainty method, where the green points are a subset of the black points. The uncertainty-based method is designed to select the points with the highest uncertainty associated with each class assignment. The regions where the class clusters overlap the most are the regions with the highest uncertainty. 

\begin{figure}[!htbp]
\includegraphics[width=0.45\textwidth]{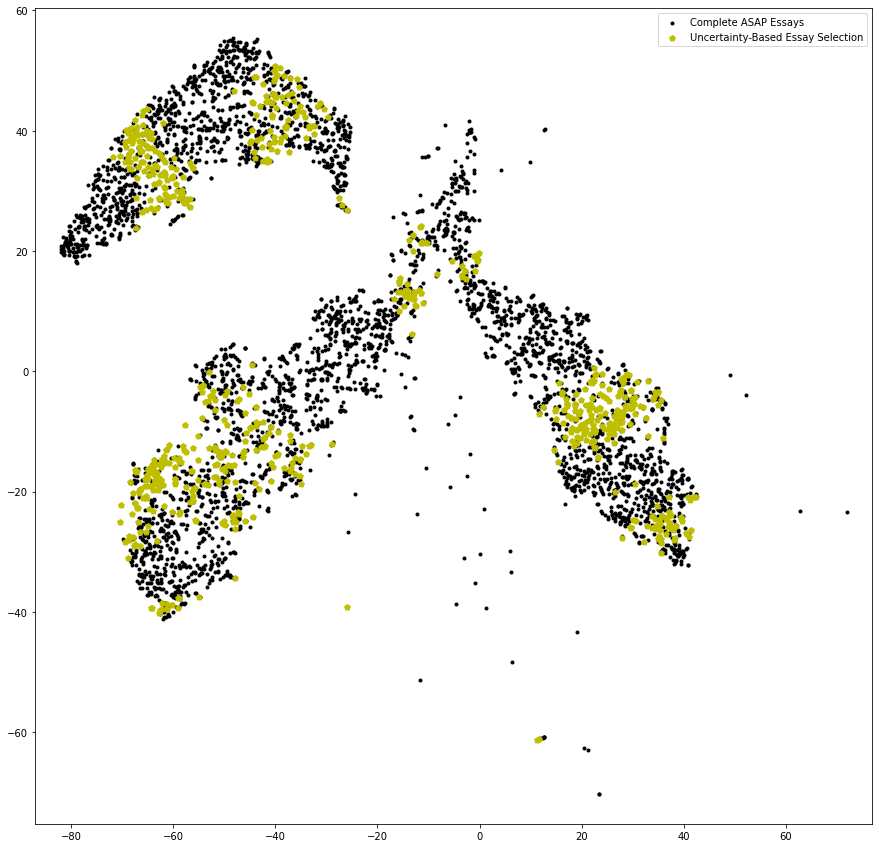}
\caption{Essays selected using uncertainty-based AL method}
\label{fig:figure_1}
\end{figure}

\subsection{Topological-Based Selection}

Topological-based AL focuses on resembling each class cluster using a minimal set of data points \citep{song2022tam}. The machine learning model learns the general shape and position of each class cluster in the feature space. The idea of shape comes from the relative distance or closeness of data points in the feature space. Hence, topological AL algorithms are formed by calculating the degree of similarity between the data points \citep{chen2021topology}.

\autoref{fig:figure_2} provides a visualization of the data points that would be selected using the topological method with the entire ASAP dataset. The black points are the essays. The green points are the essays selected for scoring using the topological method, where the green points are a subset of the black points. Similar data points have nearly identical information that must be learned by the model. As a result, including adjacent data points in the training set is counter-productive for data efficiency. In addition, by excluding similar data points from the training set, the freed data points are used to form a more distinct shape which, in turn, leads to a more accurate approximation of the class clusters. The data points are selected by setting the calculated distance as the minimum between each newly selected data point and the previously selected points. Adding essays that can be used to identify the correct shape associated with the data points increases model accuracy. Hence, essays that best define the shape of the class cluster are identified and then scored by human raters \citep{hellman2019scaling}. 

\begin{figure}[!htbp]
\includegraphics[width=0.45\textwidth]{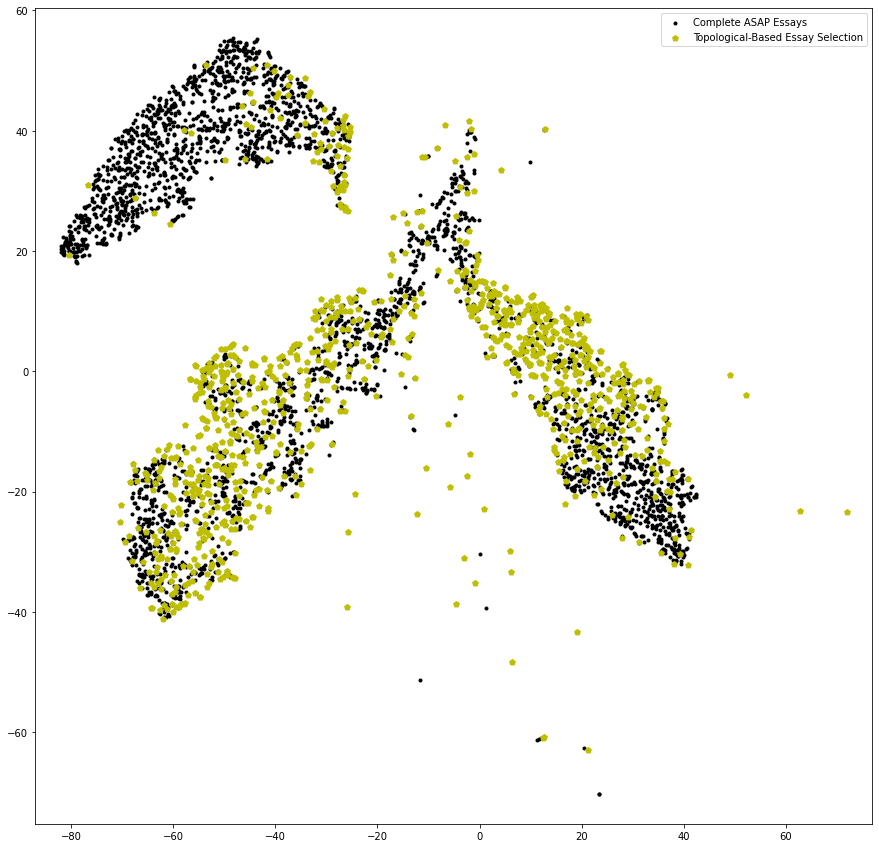}
\caption{Essays selected using topological-based AL method}
\label{fig:figure_2}
\end{figure}

\subsection{Hybrid Selection}

Hybrid selection is a new AL method created for this study. The hybrid method is based on the general shape and position of each class cluster in the feature space as with the topological approach. In addition, data efficiency may be improved by selecting data points based on the uncertainty ranking instead of a random selection of data points, as is typically the case with the topological selection method. Hence, hybrid is a combination of both the uncertainty and topological methods because shape and uncertainty are considered in selecting the essays for scoring.

\autoref{fig:figure_3} demonstrates how data points are selected by the hybrid method using the ASAP dataset. The black points are the essays. The green points are the essays selected for scoring using the hybrid method, where the green points are a subset of the black points. As with the topological method, each class cluster is resembled based on the shape using a minimal set of data points. But, unlike topological, the data points are selected by setting the calculated distance as the minimum between each newly elected data point and the previously selected points where the points with the highest uncertainty associated with each class assignment are selected.

\begin{figure}[!htbp]
\includegraphics[width=0.45\textwidth]{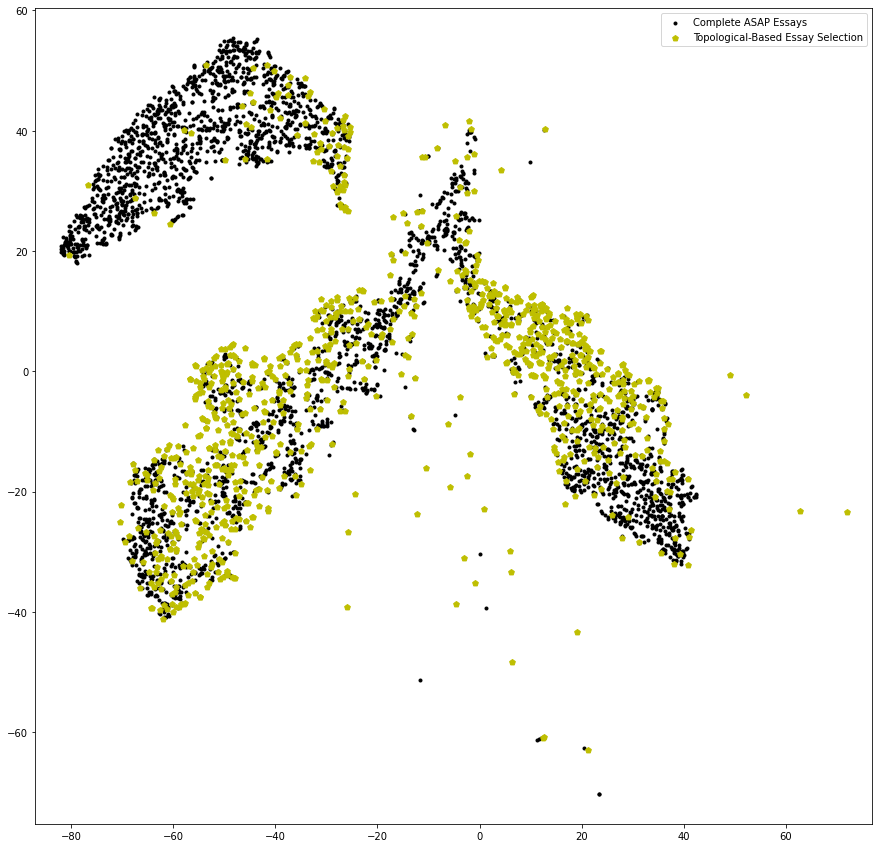}
\caption{Essays selected using the hybrid AL method}
\label{fig:figure_3}
\end{figure}

\section{Methods and Results}
\label{methods}

To demonstrate the use of AL for minimizing the number of essays that must be scored to build an accurate AES scoring model, we use the entire ASAP dataset in this study. It contained scores from human raters on 12,978 essays written using eight prompts from students in six US states at three grade levels \citep{shermis2014state}. The essays were graded by trained raters using eight different scoring rubrics. Each essay in the ASAP dataset for our study was represented by their corresponding feature vector extracted from the fine-tuned BERT model. These features contained points in 765-dimensional space, meaning each vector has 765 elements. Because the ASAP essays were scored using different rubrics on different scales, the scores were normalized to make the scores for different essays comparable. The average score for each essay was normalized between 0 and 6 using a min-max approach. This range was chosen as it is the average range of scores across the eight essays in the ASAP dataset.

The essays were then divided into two datasets. To fine tune the baseline BERT model, the first half of the essays were used as a classification dataset. The baseline model was trained on this portion to learn the text presentations by categorizing texts using their essay scores. The second half contained the new unscored essays. The data in this half served as human ratings which were only available for the selected data points by the respective algorithms.

An AL method that achieves the optimal performance with the smallest number of essays is considered to be efficient. Therefore, calculating the number of training samples needed for each method to achieve maximum accuracy demonstrated the efficiency of the method. Efficiency can be defined as the model accuracy divided by the number of training essays. We evaluated three different accuracy percentages. The lowest outcome was 85\%. This means that the accuracy achieved by the AL dataset is 85\% of the accuracy achieved when training was conducted on the full ASAP dataset. The highest outcome was 95\%. This means that accuracy achieved by the AL dataset is 95\% of the accuracy achieved when training was conducted on the full ASAP dataset. We adopted QWK as our measure of agreement because it was the official evaluation metric of the ASAP competition where the dataset of the current study originated.

\autoref{tab:percentage} shows the percentage of training essays that need to be selected by each AL method in order to achieve the target QWK. This percentage represents data efficiency, where efficiency can be defined as the model accuracy divided by the number of training essays. Because essay scoring is both time consuming and costly, achieving the target QWK using the smallest portion of selected text is desirable. The AL method that produces the target QWK using a smaller dataset is considered to be efficient. These results demonstrate two important outcomes. First, all of the methods can be used to select essays efficiently in the 85\% and 90\% conditions. To achieve a QWK that is either 85\% or 90\% accurate required less than 1\% of the original data used in the ASAP competition. Second, the topological method was the most efficient in the 95\% condition (1.8\%) and the uncertainty method the least efficient (5.1\%). The hybrid method was in the middle (4.6\%). Despite these differences, the results across all three methods were strong. That is, to achieve a QWK that is 95\% accurate, about 5\% of the data or less from the original ASAP dataset is required depending on the AL method that is used.

\begin{table}[!htbp]
\centering
\caption{Percentage of Data Required to Reach Three Levels of Accuracy as a function of Three AL Methods}
\label{tab:percentage}
\begin{tabular}{cccc}
\toprule
\textbf{Target QWK} & \textbf{Uncertainty} & \textbf{Topological} & \textbf{Hybrid} \\
\midrule
95\% & 5.1\% & 1.8\% & 4.6\% \\
90\% & <1.0\% & <1.0\% & <1.0\% \\
85\% & <1.0\% & <1.0\% & <1.0\% \\
\bottomrule
\end{tabular}
\end{table}

The growth rate accuracy across the three methods can also be evaluated. \autoref{fig:figure_4} compares the QWK achieved by each of the AL methods using different percentages of the total ASAP dataset ranging from 0.50\% to 20\%. By using different percentages of the total ASAP dataset, we can determine how AL affects a scoring budget. In other words, we can determine how accurate essay scoring is when you allocate a budget to score, for instance, 5\% of the essays compared to 15\% of the essays from the total dataset. This figure demonstrates that different percentages results in different levels of efficiency meaning that the growth rate accuracy across the three methods is not linear. For example, using 1\% of the original ASAP sample, the hybrid methods are the most efficient (74.7\%). Using 5\% of the original sample, the hybrid method is again the most efficient (76.4\%.). Using 10\% of the original ASAP sample, the uncertainty-based method is most efficient (78.0\%.). Using 15\% of the original ASAP sample, the hybrid method is most efficient (78.2\%.). Finally, using 20\% of the original ASAP sample, the hybrid method is again the most efficient (78.5\%.). These results demonstrate that the three AL methods produce different levels of efficiency under different sample size allocations.

\begin{figure}[!htbp]
\includegraphics[width=0.45\textwidth]{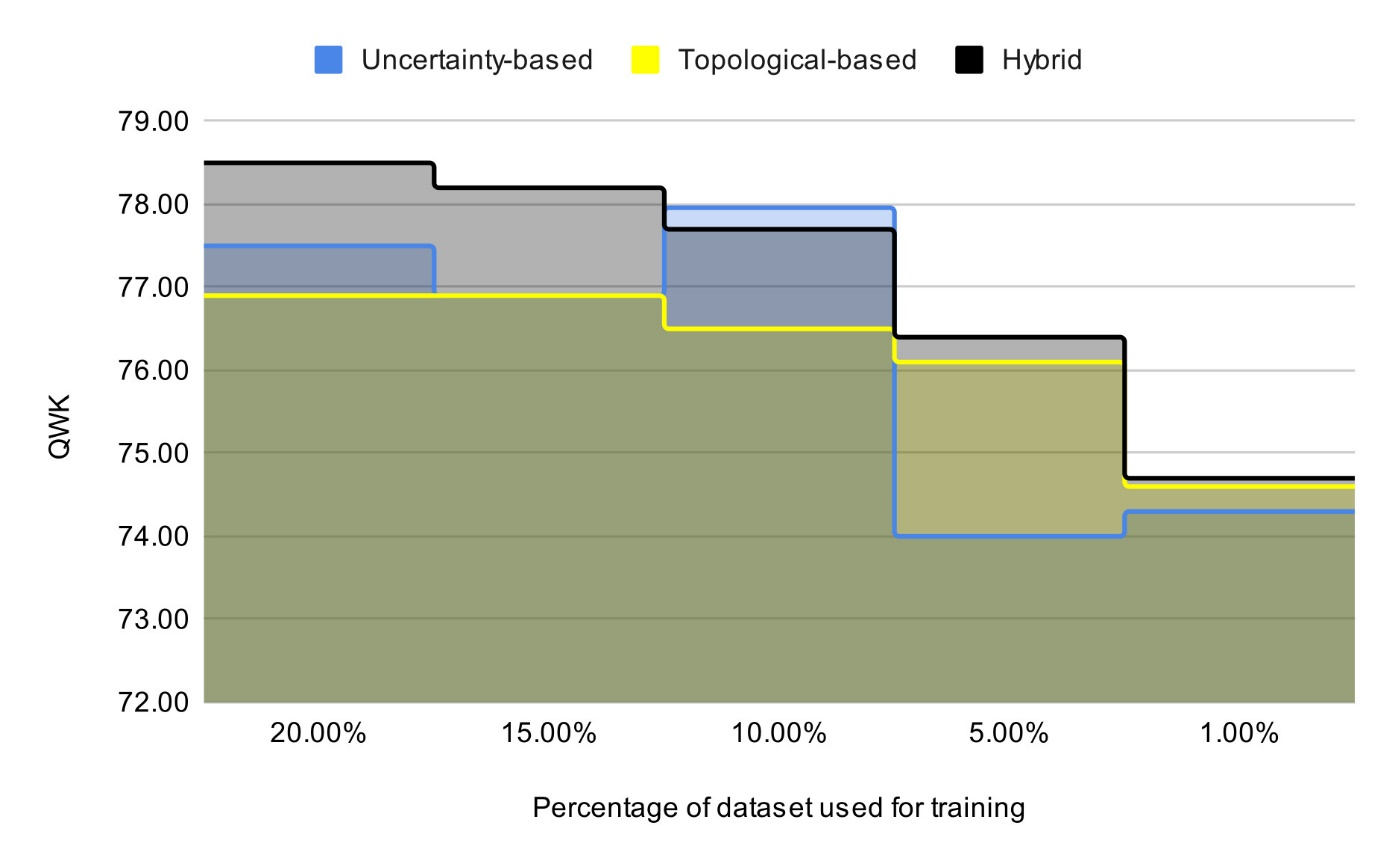}
\caption{The growth rate accuracy across the three AL methods}
\label{fig:figure_4}
\end{figure}

\section{Conclusions and Discussion}
\label{conclusions}

The purpose of this study was to describe and evaluate three AL methods than can be used to minimize the number of essays that must be scored by human raters while still providing the data needed to train a modern AES system. Research on AES has become increasing important because it serves as a scoring method for evaluating students’ written-responses at scale. Scalable methods for scoring written responses are needed as students migrate to online learning environments resulting in the need for new practices that allow educators to evaluate large numbers of written-response assessments efficiently and economically. AL is a branch of artificial intelligence that attempts to use a minimal set of scored data to build a reliable machine learning model. AL allows educators to build modern AES systems while adhering to a limited scoring budget because the number of essays that must be scored by human raters is minimized. Three AL methods were introduced and evaluated. Uncertainty-based AL identifies data points where the machine learning model has the lowest confidence associated with a classification. Topological-based AL identifies data points that best resemble each class cluster. Hybrid-based AL—a new method we created for this study—is a combination of both the uncertainty and topological methods because shape and uncertainty are considered in selecting the essays for scoring. These three AL methods were evaluated with the classified data created from a deep learning neural network language model called BERT. BERT was tuned on the ASAP dataset in order to improve its ability to score the essays. The tuned model learns the properties of texts with different essay scores and incorporates this information into the generated feature vectors containing the language context which, in our study, noticeably improved the performance of the scoring model.

An AL algorithm that yields reliable classification performance with the smallest number of essays is considered to be efficient. Because essay scoring is both time consuming and costly, achieving the target QWK using the smallest portion of selected text is desirable. We evaluated three levels of accuracy ranging from 85\ to 95\%. QWK served as our measure of agreement. All three AL methods were efficient across the three levels of accuracy. The topological method was the most efficient in the 95\% condition and the uncertainty method the least efficient. We also demonstrated that to achieve a QWK that is 95\% accurate relative to the full ASAP dataset, a mere 5\% of the data must be sampled and scored regardless of the AL method that is used. We also demonstrated that the growth rate accuracy across the three methods was not linear. The three AL methods produce different levels of efficiency under different sample size allocations. For example, using 1\%, 5\%, 15\%, and 20\% of the original ASAP sample, the hybrid method was the most efficient whereas using 10\% of the original sample, the uncertainty-based method was most efficient.

\section{Implications for Practice}
\label{implications}

The results of our study have two implications for practice. The first implication is time. We noted earlier that one important cost of using AES is that large datasets that represent each scoring category are required to train supervised machine learning models in order to ensure high agreement between the human rater and computer. In other words, conventional wisdom states that more training data are better for supervised machine learning \citep{stevenson2013shermis, williamson2012framework}. When state-of-the-art deep learning models are used, even larger samples are needed. Although there is no single answer to how much data is needed for training, the dataset should be in the thousands of responses\citep{stevenson2013shermis}. This requirement of providing the AES system with large numbers of essays can easily be addressed using an online educational platform. The challenge is scoring these essays. It is very time consuming to have human raters score a large number of essays that represent each scoring category in a short period of time. AL methods can be used to substantially minimize the number of essays that need to be scored by human raters. The strategy we implemented does not rely on securing large amounts of scored data that represents each category in the rubric. Instead, our strategy relies on identifying the essays that are either the most challenging to classify or the most representative of the general shape and position of each class cluster and then ensuring that the machine learning algorithm has access to these data. As we demonstrated, the essays that are the most challenging to classify or most representative of the class cluster actually represent a small percentage of the total sample. Using an AL strategy for selecting essays means that, first, the machine learning algorithm receives the most useful data for conducting the classification task and, second, these data represent a small percentage of the total number of essays. We demonstrated that to achieve a QWK that is 95\% accurate, 1.8\% of the training essay sample from the original ASAP analysis must be scored using essays selected using the topological method. This outcome serves as 98.2\% savings compared to the number of essays that were initially required in the ASAP analysis. Hence, the time saving is dramatic because human raters are only required to score those essays that help the machine to achieve a high level of scoring accuracy.

The second implication is cost. Cost savings are tangible because the data efficiency measure maps directly onto the number of essays that must be scored. Data efficiency is defined as the model accuracy divided by the number of training essays. Because essay scoring is costly, achieving the target QWK using the smallest portion of selected text is desirable. Growth rate accuracy can be used as a direct measure of cost because different percentages of the total ASAP dataset can be scored. Educators can determine the accuracy of their essay scoring when, for instance, 5\% of the essays are scored compared to 20\% of the essays. The results in Figure 4 demonstrate that the three AL methods produce different, albeit all consistently high, levels of efficiency under different sample size allocations. Consequently, practitioners can reduce their scoring budget dramatically because only a small percentage of the essays in the full sample are required to create a state-of-the art scoring model. The decreased number of essays that must be scored can be translated into a cost savings for practitioners because their scoring budget can be reduced. We demonstrated that the magnitude of the scoring budget reduction differed by AL method.

In addition, cost saving can be realized when transformers are used for feature extraction because of the ease of use and the generalizability of transformer models. Transformer models are now used extensively resulting in the formation of large supporting communities of users. As a result, transformer language models are accessible and have easy-to-use programming interfaces. In addition, transformers are available for more than 200 different languages. This language diversity makes transformer-based methods easily applicable to a large set of languages with minor modifications which means the methods described in our study can easily be applied to essays written in hundreds of different languages. As a result, the AL methods described in this study can be used to substantially minimize the number of essays that must be scored by human raters while still providing the data needed to efficiently train a modern AES system across hundreds of different language groups.

}

\bibliographystyle{IEEEtranN}
\bibliography{sample-bibliography}

\end{document}